\documentclass[conference]{IEEEtran}
\ifCLASSINFOpdf
\else
\fi
\usepackage{amsmath}
\IEEEoverridecommandlockouts

\usepackage{mathtools}
\usepackage{caption}
\usepackage{subcaption}
\usepackage{comment}
\usepackage{amssymb,amsthm}
\usepackage{xcolor}
\usepackage{multirow}
\usepackage{adjustbox}
\usepackage{bm}

\theoremstyle{remark}

\def\sb{{\boldsymbol\sigma}}

\def\mub{{\boldsymbol\mu}}
\def\etab{{\boldsymbol\eta}}

\def\zb{{\boldsymbol\zeta}}
\def\z{{\bm{z}}}
\def\w{{\omega}}

\def\W{{\bm{W}}}
\def\R{{\mathbb{R}}}
\def\P{{\mathbb{P}}}
\def\E{{\mathbb{E}}}
\def\d{{\partial}}
\def\D{{\boldsymbol{\mathcal{D}}}}

\begin{document}
%
\title{Learning the Evolution of Correlated Stochastic Power System Dynamics}

\author{
\IEEEauthorblockN{Tyler E. Maltba \IEEEauthorrefmark{1}\IEEEauthorrefmark{2},
Vishwas Rao\IEEEauthorrefmark{1},
Daniel Adrian Maldonado\IEEEauthorrefmark{1}}
\IEEEauthorblockA{\IEEEauthorrefmark{1} Mathematics and Computer Science Division, Argonne National Laboratory,
Lemont, IL, USA}
\IEEEauthorblockA{\IEEEauthorrefmark{2} Department of Statistics, University of California, Berkeley, Berkeley, CA, USA}
\thanks{This material was based upon work supported by the U.S. Department of Energy (DOE), Office of Science,
Office of Advanced Scientific Computing Research (ASCR) under Contract DE-AC02-06CH11347. We acknowledge the support from US DOE office of Electricity Delivery and Energy Reliability – Advanced Grid Modeling (AGM) Program. We also acknowledge
partial support by the US National Science Foundation (NSF) through grants DGE 1752814 to Tyler E. Maltba.}}


%


\maketitle

\begin{abstract}
A machine learning technique is proposed for quantifying uncertainty in power system dynamics with spatiotemporally correlated stochastic forcing. We learn one-dimensional linear partial differential equations for the probability density functions of real-valued quantities of interest. The method is suitable for high-dimensional systems and helps to alleviate the curse of dimensionality. 
\end{abstract}


\section{Introduction}\label{sec:intro}
To reduce carbon emissions, electrical power systems are increasingly incorporating renewable generation resources into the energy mix. These resources are often dependent on weather inputs and, as a result, they behave stochastically in the short and long terms, posing planning and operational challenges. Including stochastic fluctuations in existing power system models is technically challenging, and recent efforts have been made to introduce probabilistic analysis techniques in power flow and dynamic simulation \cite{Milanovi2017}.

In this work, we are concerned with quantifying the uncertainty in dynamics when the model is forced by a stochastic process. Milano and Z\'arate-Mi\~nano \cite{milano2013} introduced a new formulation to model the differential-algebraic dynamic equation (DAE) as a stochastic-algebraic differential equation (SDAE) and used Ohrnstein-Uhlenbeck (OU) processes to represent the stochastic fluctuations in the power grid. The OU process was also determined in  \cite{Roberts2016} to be a good candidate for modeling stochastic fluctuations of the load in the short term when measured with a Phasor Measurement Unit. Moreover, OU processes were recently used to study the impact of correlated stochastic processes in dynamic performance \cite{Adeen2021}.

While these studies highlight the importance of quantifying the effect of stochasticity in power system dynamics, this task is computationally demanding. Stochastic trajectories can be simulated using standard numerical techniques such as Euler-Maruyama \cite{Wang2011}. However, complete characterization of the state probability distribution function (PDF) over time is technically challenging. Due to the curse of dimensionality, standard (multilevel) Monte Carlo (MC) methods cannot be used to fully characterize the PDF except for low-dimensional systems. Characterization of the PDF evolution is sometimes possible by fully (numerically) integrating the Fokker-Planck equation (FPE) \cite{Wang2013bis}, but is also limited to low-dimensional systems. Moreover, other methods such as polynomial chaos and stochastic finite elements (e.g., \cite{Qiu2021}) often do not provide any computational speed-up if the system's noise is characterized by a short temporal correlation length (see the references in \cite{Wang2013}).

Outside of the power systems community, novel machine learning techniques for partial differential equations (PDEs) have been used to efficiently learn evolution equations for PDFs of system states. We refer to such equations as PDF equations, and unlike the FPE \cite{Risken1996}, many are unclosed. Particularly for low-dimensional systems, there has been recent success in learning PDEs for the full joint PDF of all system states \cite{Maltba2021}. However, since many power systems are high-dimensional, there is a need for new methodologies. For the purpose of forward uncertainty quantification, it is often sufficient to study low-dimensional, or even scalar quantities of a high-dimensional system. 

In this paper, we develop a data-driven method to learn reduced-order PDF (RO-PDF) equations for the PDFs of real-valued quantities of interest {(\sc{QoI})}. The resulting PDEs are linear, one-dimensional, and easily solvable with standard numerical schemes. We show that the unknown terms can be learned from relatively few\footnote{This terminology refers to the number of simulations required by the method being only a small fraction of those needed for a fully converged Monte Carlo method.} MC simulations of the underlying high-dimensional system, meaning that the method is scalable and does not suffer from the curse of dimensionality to the same extent as traditional methods. 

In Section~\ref{sec:model}, we introduce a stochastic power model and discuss how to account for correlation between the driving stochastic (noise) processes in the model. The methodology for learning RO-PDF equations for real-valued {\sc{QoI}} in the stochastic power model is discussed in Section~\ref{sec:methods}. Accuracy and scalability of the method is successfully tested on the multi-machine classical model with stochastic power injections modeled by correlated OU processes in Section~\ref{sec:studies}.


\section{Stochastic Dynamics Model}\label{sec:model}

In this work we are concerned with quantifying the uncertainty in power system dynamics. This is often achieved by integrating a semi-explicit index-1 DAE system
\begin{align}\label{dae}
\begin{split}
    \dot{\bm{x}} &= g(\bm{x}, \bm{y}, t) \,, \\
    0 &= h(\bm{x}, \bm{y}, t) \,.
\end{split}
\end{align}
To account for time-varying, uncertain processes such as renewable generation or load injections, it is possible to augment this system with a stochastic differential equation (SDE) \cite{milano2013}, which we call the noise model. We use an $n$-dimensional It\^o SDE driven by independent Wiener processes \cite{Adeen2021}. They have the general form 
\begin{align}\label{noise1}
d{\etab}(t)={\bm{a}}({\etab}(t))dt+{\boldsymbol{b}}({\etab}(t))\odot{d{\bm{w}}(t)},
\end{align}
where $\bm{a}:\R^n\rightarrow\R^n$ and $\bm{b}:\R^n\rightarrow\R^n$ are respectively the drift and diffusion terms of the SDE, and $\bm{w}\in\R^n$ is a standard $n$-dimensional Wiener processes. In addition, $\bm{w}$ is a zero-mean Gaussian process with stationary, independent Gaussian increments and continuous paths. The $\odot$ notation represents the element-wise Hadamard product. The $n$-dimensional OU process is a commonly used noise model taking the form of~\eqref{noise1}.

Following \cite{Adeen2021} we generalize the noise model~\eqref{noise1} to be driven by a Gaussian process $\bm{v}\in\R^n$ (instead of $\bm{w}$) whose components are correlated. Such a process $\bm{v}$ can be constructed as a linear combination of $n$ uncorrelated Wiener processes--it has a symmetric correlation matrix $\bm{R}\in\R^{n\times n}$ with elements $R_{ij}$. 
The increments of $\bm{v}$ can be written in terms of the increments of $\bm{w}$ (see \cite{Dipple2020} and the reference therein) as
\[ d\bm{v} = \bm{C}d\bm{w}, \]
where $\bm{C}\in\R^{n\times n}$ such that $\bm{R} = \bm{C}\bm{C}^\top$. In general, $\bm{R}$ can be a stationary stochastic process; however, we restrict $\bm{R}$ to be constant and positive-definite for computational efficiency. Therefore, the Cholesky decomposition of $\bm{R}$ is unique, i.e.,  the (constant) lower triangular $\mathbf{C}$ is unique. 

The noise model~\eqref{noise1} generalized to be driven by the process $\bm{v}$ can still be written it terms of $\bm{w}$. We have
\begin{align}\label{noise2}
    d{\etab}(t)&={\bm{a}}({\etab}(t))dt+{\boldsymbol{b}}({\etab}(t))\odot{d{\bm{v}}(t)} \notag \\
    &={\bm{a}}({\etab}(t))dt+{\boldsymbol{b}}({\etab}(t))\odot\bm{C}{d{\bm{w}}(t)}.
\end{align}
Introducing $\bm{\eta}$ in~\eqref{noise2} to the differential part of \eqref{dae}, the model can be rewritten as an It\^o SDE of higher dimension, taking the general form 
\begin{align} \label{gen_sde}
    d\zb(t) &= \mub(\zb(t),t)dt + \sb(\zb(t),t)d\W(t), \notag\\
    \zb(0) &= \zb^0.
\end{align}
The solution $\zb$ to~\eqref{gen_sde} is an $\R^N$-valued ($N>n$) stochastic process whose components are made up of the components of $\bm{x}$, $\bm{y}$, and $\etab$. The initial condition $\zb^0$ may also be random, i.e., an $N$-dimensional random vector with joint PDF $f_{\zb^0}(\z) : \R^N \to \R^+$. The driving process $\W$ of \eqref{gen_sde} is a standard $M$-dimensional Wiener process. The (random) initial state $\zb^0$, drift $\mub: \R^N \to \R^N$, and diffusion $\sb: \R^N \to \R^{N\times M}$ satisfy conditions guaranteeing the existence and uniqueness of a strong solution to~\eqref{gen_sde} \cite{Oksendal2003}.

\section{Methodology}\label{sec:methods}

Let $\z\in\R^N$ be a variable in the phase space of~\eqref{gen_sde}. At any given time $t$, the state of the system is (partially) characterized by the single-point (in time) joint cumulative distribution function (CDF) $F_\zb(\z;t) = \P[\zb(t) \le \z]$ and its associated single-point PDF $f_\zb(\z;t)$. It is well-known that $f_\zb(\z;t)$ satisfies the deterministic $N$-dimensional linear PDE called the FPE, given by
\begin{align} \label{fpe}
    &\frac{\d f_\zb}{\d t} + \sum_{i=1}^N \frac{\d}{\d z_i}(\mu_i(\z,t) f_\zb) = \sum_{i=1}^N\sum_{j=1}^{M} \frac{\d^2}{\d z_i\d z_j}(\mathcal{D}_{ij}(\z,t)f_\zb) \notag\\
    &f_{\zb}(\z;0) = f_{\zb^0}(\z),
\end{align}
where $\D = \frac12 \sb\sb^\top$. 

In general, FPEs do not admit closed-form analytic solutions, and \eqref{fpe} must be solved numerically. Hence, when the dimension $N$ is large, \eqref{fpe} is a high-dimensional PDE. While there have been recent advancements in numerical solutions to high-dimensional PDEs, such as tensor train methods (see \cite{Boelens2018} among others), they are not capable of handling systems whose dimension is in the thousands, as is common in power systems. Hence, numerically solving the FPE can be intractable. However, for the purpose of forward uncertainty propagation, it is often sufficient to consider low-dimensional {\sc{QoI}}. A low-dimensional PDE for the PDF of that quantity (i.e., a RO-PDF equation) can then be learned and efficiently solved with standard numerical methods. We focus on real-valued {\sc{QoI}}. 

Consider a stochastic process (for the {\sc{QoI}}) $U(\zb(t))\in\R$, where $U: \R^N \to \R$ is a phase space function guaranteeing the existence of single-point PDF $f_U$ of $U(\zb(t))$ for all $t\ge 0$. Let $u\in\R$ be a phase space variable for $U(\zb(t))$. Then $f_U(u;t)$ is found by integrating $f_\zb(\z;t)$ against the Dirac delta function $\delta(\cdot)$:
\begin{align*}
    f_U(u;t) = \int_{\R^N} \delta(u - U(\z))f_\zb(\z;t)d\z.
\end{align*}
Multiplying the FPE \eqref{fpe} by $\delta(u - U(\z))$ and integrating with respect to $\z$ gives 
\begin{flalign}\label{ro-pdf}
    &\frac{\d f_U}{\d t} + \sum_{i=1}^N \int_{\R^N} \delta(u - U(\z))\frac{\d}{\d z_i}(\mu_i(\z,t) f_\zb)d\z &&\notag\\ &\quad\quad= \sum_{i=1}^N\sum_{j=1}^{N} \int_{\R^N}\delta(u - U(\z))\frac{\d^2}{\d z_i\d z_j}(\mathcal{D}_{ij}(\z,t)f_\zb)d\z, &&\notag\\
    &f_U(u;0) = \int_{\R^N}\delta(u - U(\z))f_{\zb^0}(\z)d\z.
\end{flalign}
This equation for $f_U(u;t)$ is unclosed since there are terms depending on the generally unknown $f_\zb(\z;t)$, and not on $f_U(u;t)$ alone.

We focus on the case where the {\sc{QoI}} are the scalar components of $\zb$ (i.e., when $U(\zb) = \zeta_j$ for $j=1,\hdots,N$). By assuming $f_\zb(\z;t)$ and its partial derivatives decay fast enough at $\pm\infty$ when the phase space is unbounded, equation \eqref{ro-pdf} can be simplified. Letting $\z_{-j} = [z_1,\hdots,z_{j-1},z_{j+1},\hdots,z_N]^\top$, the unclosed marginal PDF equation for $f_{\zeta_j}(z_j;t)$ is
\begin{align}\label{ro-pdf1}
    &\frac{\d f_{\zeta_j}}{\d t} + \int_{\R^{N-1}} \frac{\d}{\d z_j}(\mu_j(\z,t)f_\zb)d\z_{-j}  \notag\\
    &\quad\quad\quad\quad\quad\quad\quad\quad\quad\quad= \int_{\R^{N-1}} \frac{\d^2}{\d z_j^2}(\mathcal{D}_{jj}(\z,t)f_\zb)d\z_{-j},\notag\\
    &f_{\zeta_j}(z_j;0) = \int_{\R^{N-1}} f_{\zb^0}(\z)d\z_{-j}.
\end{align}
Writing the joint PDF $f_{\zb}$ as the product of the marginal and conditional PDFs (i.e., $f_\zb \equiv f_{\zeta_j}f_{\zb_{-j}|\zeta_j}$), the integrals in \eqref{ro-pdf1} are conditional expectations, yielding
\begin{align}\label{ro-pdf2}
    \frac{\d f_{\zeta_j}}{\d t} &+  \frac{\d}{\d z_j}\Big(\E\big[\mu_j(\zb(t))\,\big|\,\zeta_j(t)=z_j\big]f_{\zeta_j}\Big)\notag\\
    &= \frac{\d^2}{\d z_j^2}\Big(\E\big[\mathcal{D}_{jj}(\zb(t))\,\big|\,\zeta_j(t)=z_j\big]f_{\zeta_j}\Big)
\end{align}
with vanishing (or zero-flux) boundary conditions. If $\mu_j(\cdot)$ and $\mathcal{D}_{jj}(\cdot)$ are (at least partially) separable in $z_j$, then the marginal PDF equation \eqref{ro-pdf2} can be further simplified, i.e., made to be more physics-informed. Typical expansion-based closures of \eqref{ro-pdf2} such as Kubo's cumulant expansion \cite{Kubo1962} are computationally intractable. Hence, there is a need for direct data-driven estimation of the conditional expectations. We learn such conditional expectations, which we call regression functions, from relatively few MC realizations of \eqref{gen_sde}.

To solve~\eqref{ro-pdf2}, one must choose a numerical scheme such as a finite difference, volume, or element method, and a computational mesh/grid. We consider uniform spatial meshes with domains large enough to ensure that the PDE solutions do not interact with the computational boundaries. These boundaries are chosen based on a small number of SDE sample paths. We then approximate the vanishing boundary conditions with homogeneous Dirichlet conditions with negligible effects.

The cell size $\Delta z_j$ of the spatial mesh can be manually set \emph{a priori}, or treated as a hyper-parameter to be tuned to balance computational efficiency and  truncation error of the discretization. However, like the mesh boundaries, we choose $\Delta z_j$ based on sample paths. In fact, we have found appropriate cell sizes can be computed via rule-of-thumb bandwidth estimators typically used for kernel density estimation (KDE). 

Once a spatial grid is chosen, regression functions in~\eqref{ro-pdf2} are estimated while numerically time stepping the PDE. At discrete times, the regression functions are treated as phase space functions and estimated at the discrete spatial locations required by the PDE discretization. 

For some SDE systems, including some power system models, the regression functions may vary nonlinearly in time, but remain linear in space. In such instances, standard linear regression methods can be employed with little costs. When parametric regression is inappropriate, several nonparametric methods such as moving averages, splines, local polynomial regression, neural networks, etc., can be employed \cite{Hastie2009}. Out of these, the first two were successively used in \cite{Brennan2018} for nonlinear ODEs with random initial conditions. However, we found Gaussian local linear regression (LLR) to have a more desirable trade-off between costs and accuracy. LLR has only a single hyper-parameter (the kernel bandwidth), which can be efficiently estimated with standard cross validation (CV) methods such as k-fold CV \cite{Hastie2009}. Although beyond the scope of this paper, and similar to the approach in \cite{Brennan2018}, it is worth noting that information content of the data used for regression can be measured by deriving a new system of PDEs equivalent to \eqref{ro-pdf2}. Solving the system with the same data-driven approach and comparing its solution to that of \eqref{ro-pdf2} can determine if more useful data is needed.


\section{Case studies}\label{sec:studies}

We consider the \emph{multi-machine classical model} \cite{anderson_power_2003}, chosen for its simplicity, which allows the scalability of the RO-PDF method to be readily analyzed. Here, we consider stochastic power injections and write the complete system:
\begin{align}
    d\hat v_i(t) &= 0 &&\notag\\
    d\w_i(t) &= \frac{\w_R}{2H_i}\Big[-D_i(\w_i(t)-\w_R) + P_i&&  \notag\\
    &- \hat v_i\sum_{j=1}^n \hat v_j\Big(g_{ij}\cos(\delta_i(t)-\delta_j(t)) &&\notag\\  &+b_{ij}\sin(\delta_i(t)-\delta_j(t))\Big) + \eta_i(t)\Big]dt && \notag \\
    d\delta_i(t)&= (\w_i(t) - \w_R)dt  && \notag\\
    d\etab(t) &= -\theta\etab(t)\,dt + \alpha\sqrt{2\theta}\bm{C}\, d\bm{w}(t) \label{mmc_modf}
\end{align}
for $i = 1,...,n$, where $\w_i$ and $\delta_i$ are respectively the speed and angle of machine $i$, and $\etab$ is the driving OU noise correlated across machines according to the constant matrix $\bm{R} = \bm{C}\bm{C}^\top$. In general, the drift and diffusion of $\etab\in\R^n$ may vary across generators, but we take them to be constants $\theta = 1$ and $\alpha = 0.05$ for simplicity. $\etab$ is initialized with the Gaussian random vector $\etab(0) \sim \mathcal{N}(\mathbf{0},\alpha^2\mathbf{R})$.
We initialize the system around an equilibrium point by solving the deterministic power flow with Matpower \cite{Zimmerman2011}. We then treat each voltage magnitude $\hat v_i$ of machine $i$ as a folded Gaussian random parameter by taking $\hat v_i(0) \sim |\mathcal{N}(v_i,(0.05\bar v)^2)|$, where $v_i$ is the constant equilibrium voltage of machine $i$ and $\bar v = \frac{1}{n}\sum_{i=1}^n v_i$ is the average over all equilibrium voltages. The constant $\w_R$ is the equilibrium speed of the system, $g$ and $b$ are constant admittance matrices, $P_i$'s are constant equilibrium total power bus injections, $H_i$'s are constant inertia coefficients, and $D_i$'s are constant damping factors. In all of the experiments, we set $w_R = H_i = D_i = 1$ for all $i$ and simulate~\eqref{mmc_modf} up to the final time $T_f = 10$.

The initial speeds and angles of~\eqref{mmc_modf} are treated as random variables and samples of their distributions are computed by burning in MC simulations of~\eqref{mmc_modf} in the following manner. We first take $\w_i(0) = w_R$ and $\delta_i(0)$ as the equilibrium points from solving the deterministic power flow equations with Matpower. The  SDE~\eqref{mmc_modf}, with random $\hat v_i$ and $\etab$ included, is then simulated until time $T_f$. The resulting realizations of $\w_i$ and $\delta_i$ at time $T_f$ are treated as samples of the random variables $\w_i(0)$ and $\delta_i(0)$. Hence, we can think of~\eqref{mmc_modf} as being initialized close to equilibrium, albeit stochastic. In our experiments, we consider three test cases: WSCC 9-Bus System, IEEE 30-Bus System, and IEEE 57-Bus System, which from here on will be referred to as Case 9, Case 30, and Case 57, respectively.

\subsection{Covariance model}\label{sec::cov}
Often, the correlation of the stochastic process is spatial in nature \cite{Lenzi2021}. Thus, it is natural to use spatial kernels to determine the covariance model. We consider three choices for the correlation matrix $\bm{R}= \bm{C}\bm{C}^\top$ of the noise $\etab$ in~\eqref{mmc_modf}. We denote the different choices of $\bm{R}$ by $\bm{R}^U$ for uncorrelated, $\bm{R}^E$ for exponentially correlated, and $\bm{R}^C$ for constant correlation. In the uncorrelated cases, $\bm{R}^U$ is the identity matrix. For constant correlation, for all connected buses $i\neq j$, we set $R^C_{ij}=R^C_{ji}=0.44$ for Case 9, and $R^C_{ij}=R^C_{ji}=0.36$ for Case 30 and Case 57. These values represent a low-to-moderate level of correlation.

For the exponentially correlated case $\bm{R}^E$, we first approximate the distance $d_{ij}\in\R^+$ in miles between two connected buses $i$ and $j$ by $d_{ij} = \sqrt{X_{ij}B_{ij}}$, where $X_{ij}$ and $B_{ij}$ are respectively the reactance and susceptance of line ${i,j}$. We replace zero susceptances with the minimum of all non-zero susceptances to ensure that $d_{ij}>0$ for connected machines. We then take 
\begin{align*}
    R^E_{ij} = R^E_{ji} = \exp\left(-\frac{d_{ij}}{\lambda}\right)
\end{align*}
for connected buses $i\neq j$. We take the scaling factor $\lambda = 82$ for Case 9, $\lambda = 14.5$ for Case 30, and $\lambda = 5$ for Case 57. These values were chosen to be the (approximate) maximum $\lambda$ values such that $\bm{R}^E$ remains positive-definite. 

\subsection{Reduced-order PDF equations}
Regarding the SDE model~\eqref{mmc_modf}, we consider the speeds $\w_i(t)$ and angles $\delta_i(t)$ of each machine as the real-valued {\sc{QoI}}. In both cases, equation~\eqref{ro-pdf2} reduces to a conservative advection equation since their corresponding right-hand-sides of~\eqref{mmc_modf} lack diffusion. To demonstrate, let $z\in \R$ be a phase space variable for $\w_i(t)$. Then the unclosed RO-PDF equation for the marginal PDF $f_{\w_i}(z;t)$ of $\w_i(t)$ can be written as 
\begin{align}\label{ro_mmf1}
    &\frac{\d f_{\w_i}}{\d t} +\frac{\d}{\d z}\Big[\Big(\frac{\w_R}{2H_i}\Big)\big(-D_i(z-\w_R) + P_i -m(z;t)\big)f_{\w_i}\Big] \notag\\ &= 0,
\end{align}
where the regression function is given by
\begin{flalign}\label{ro_mmf_coe}
    m(z;t) &= \E\Big[\hat v_i\sum_{j=1}^n \hat v_j\Big(g_{ij}\cos(\delta_i(t)-\delta_j(t)) &&\notag\\  &+b_{ij}\sin(\delta_i(t)-\delta_j(t)\Big) - \eta_i(t) \Big|\, \w_i(t) = z\Big].
\end{flalign}
The analogous equation for the angle $\delta_i(t)$ is simply
\begin{align}\label{ro_mmf2}
    \frac{\d f_{\delta_i}}{\d t} +\frac{\d}{\d z}\Big[\big(\E[\w_i(t)\,|\,\delta_i(t) = z] - \w_R\big)f_{\delta_i}\Big]= 0.
\end{align}

\subsection{Experiments}
We investigate the accuracy and scalability of our method for systems both in and out of ``equilibrium''. The latter is achieved by introducing a single line failure for each of the three test cases. In our experiments we remove line 8-9 for Case 9, line 6-8 for Case 30, and line 36-37 for Case 57.

The initial conditions for~\eqref{ro_mmf1} and~\eqref{ro_mmf2} are computed via MC simulations as described in the paragraphs following equation~\eqref{mmc_modf}. These realizations are then post-processed with Gaussian KDE. The number of MC simulations used to compute the initial conditions is restricted to be no more than the maximum number of those needed to learn the regression functions. In non-equilibrium, i.e., when a line failure is present, we use the same initial conditions for~\eqref{ro_mmf1} and~\eqref{ro_mmf2} as those used when no line failure is present. 

\begin{figure}[h!] 
\centering  
\includegraphics[width=.45\linewidth]{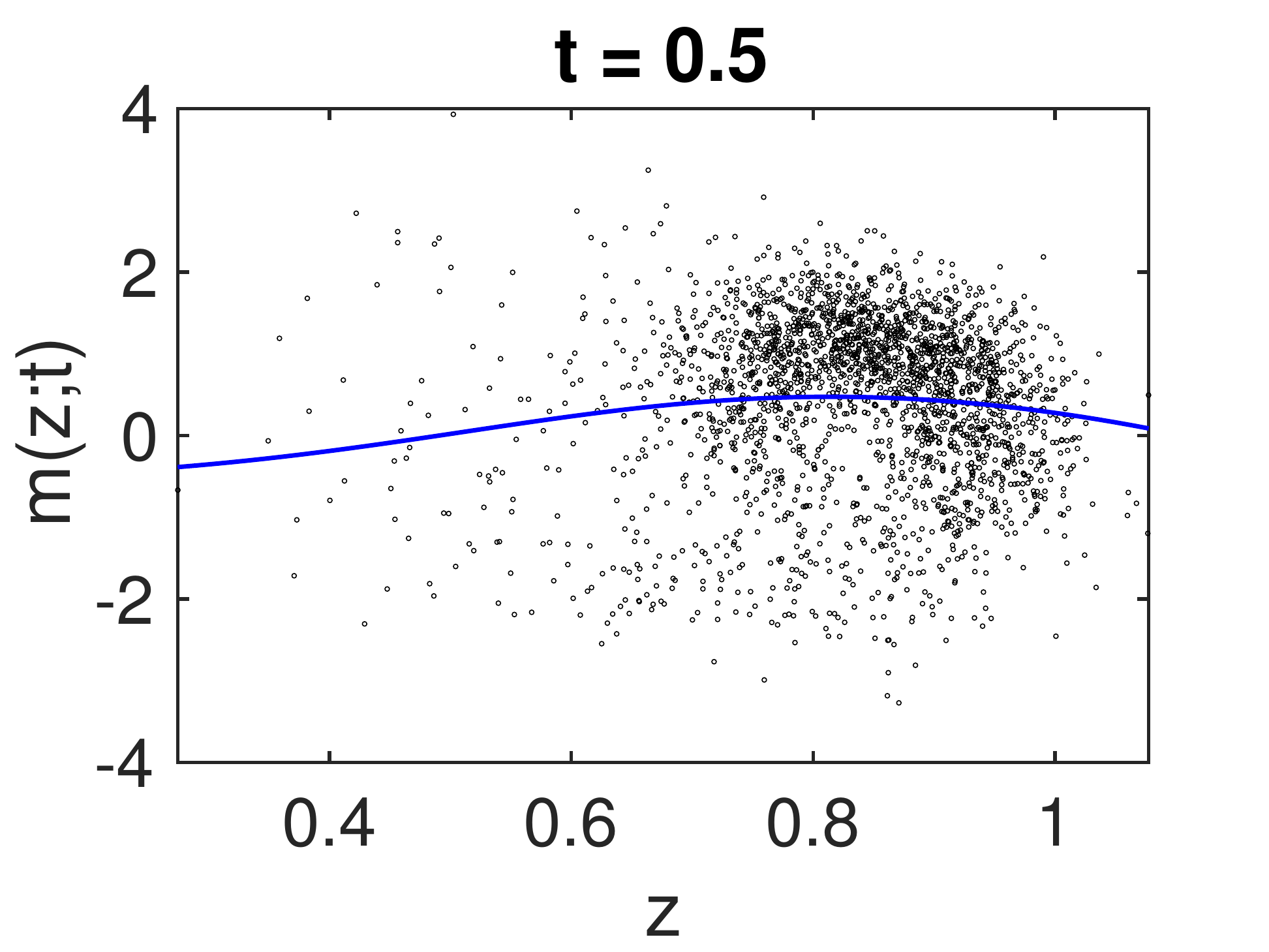}
\includegraphics[width=.45\linewidth]{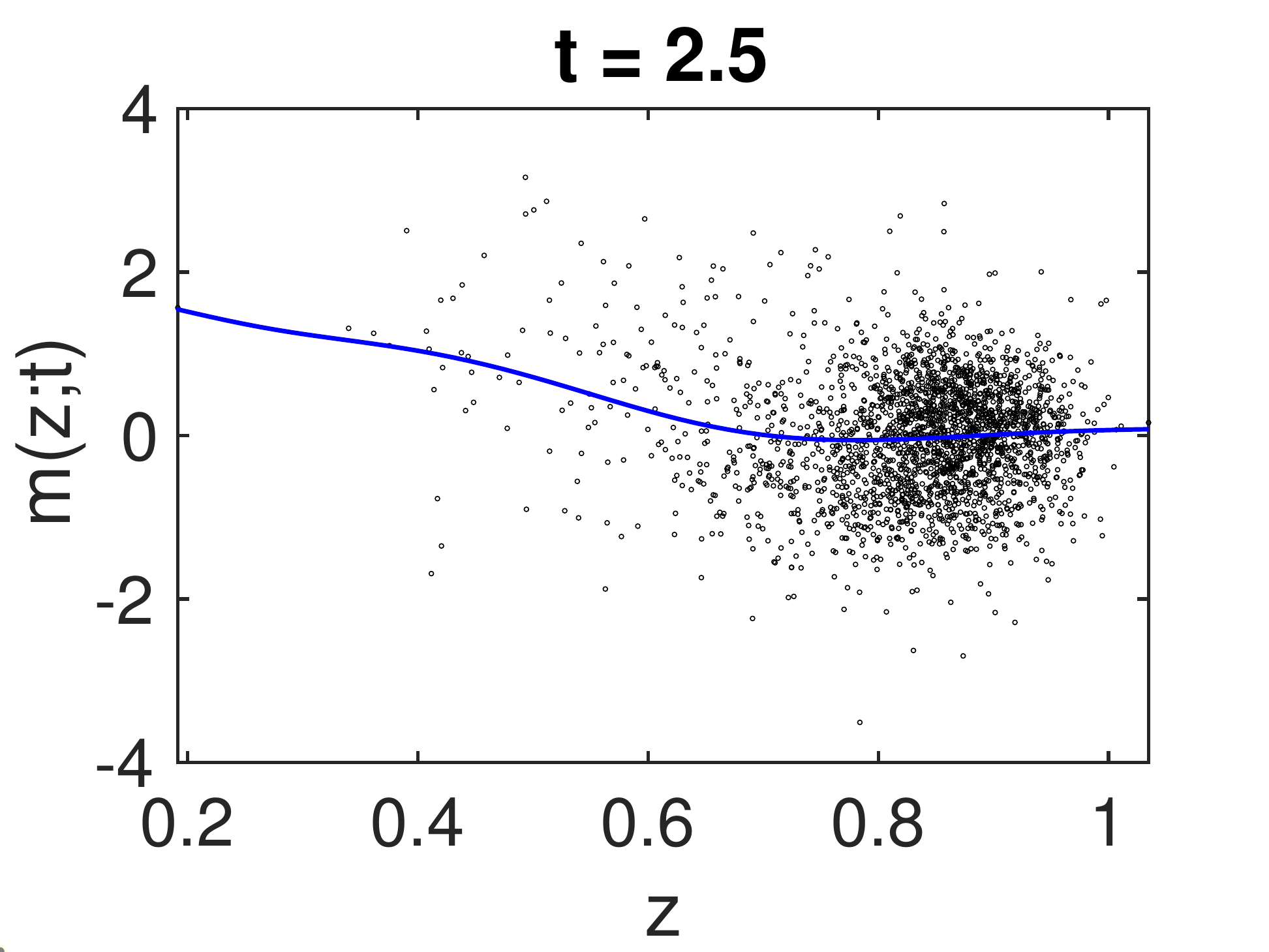}\\
\includegraphics[width=.45\linewidth]{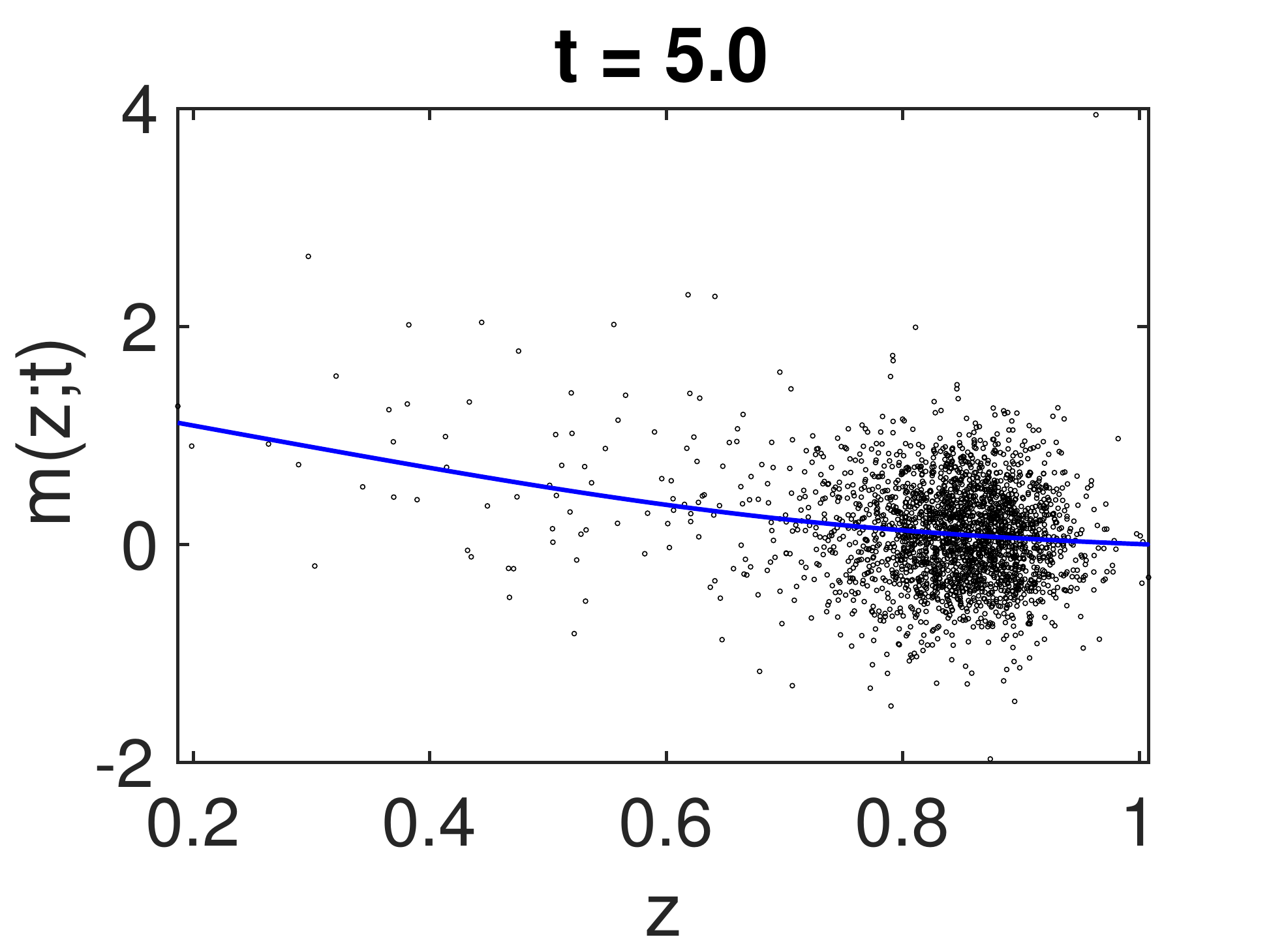}
\includegraphics[width=.45\linewidth]{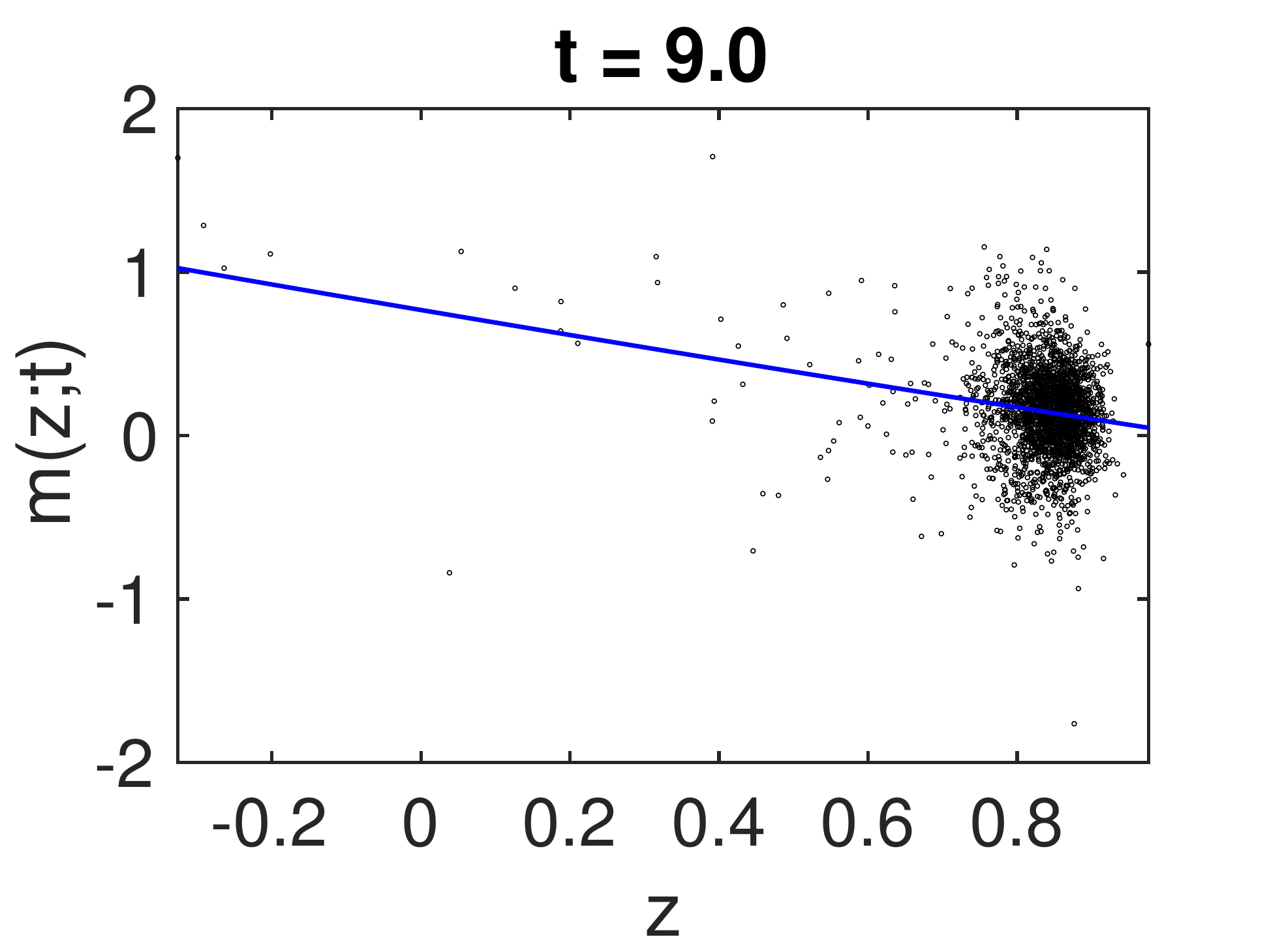}
\caption{2000 MC realizations of the speed $\w_4$ at times $t=0.5$, $2.5$, $5$, and $9$, in Case 57 with a line failure and $\bm{R}=\bm{R}^C$. The blue curve is the regression function in~\eqref{ro_mmf_coe} estimated by Gaussian LLR with 10-fold CV.}
\label{fig:coeff}
\end{figure}

When~\eqref{mmc_modf} is in equilibrium, i.e., no line failure, the regression functions for both the speed and angle RO-PDF equations are linear for all observed $t>0$; therefore, linear regression is used. This is also true for the angles even when a line failure is present. In non-equilibrium, the regression functions for the speeds are nonlinear at early times; hence, we use Gaussian LLR with 10-fold CV for kernel bandwidth selection. After some time, the systems transition to a new equilibrium and the regression functions become linear. We manually select the time at which the regression is switched from LLR to linear regression; however, efficient hypothesis testing~\cite{Hastie2009} can be used to automatically select the transition time. Fig.~\ref{fig:coeff} gives the learned regression function $m(z;t)$ in~\eqref{ro_mmf_coe} at various observed times for $\w_4(t)$ in Case 57 with a line failure and $\bm{R}=\bm{R}^C$. Notice how $m(z;t)$ becomes linear as time progresses. By time $T_f=10$, $m(z;t)$ is nearly constant in $t$ and linear in $z$ such that the advection coefficient in~\eqref{ro_mmf1} is approximately zero. Hence, the marginal PDF solution $f_{\w_4}(z;t)$ is close to steady-state. 

\begin{figure}[h]
	\centering
	\includegraphics[width=0.47\textwidth]{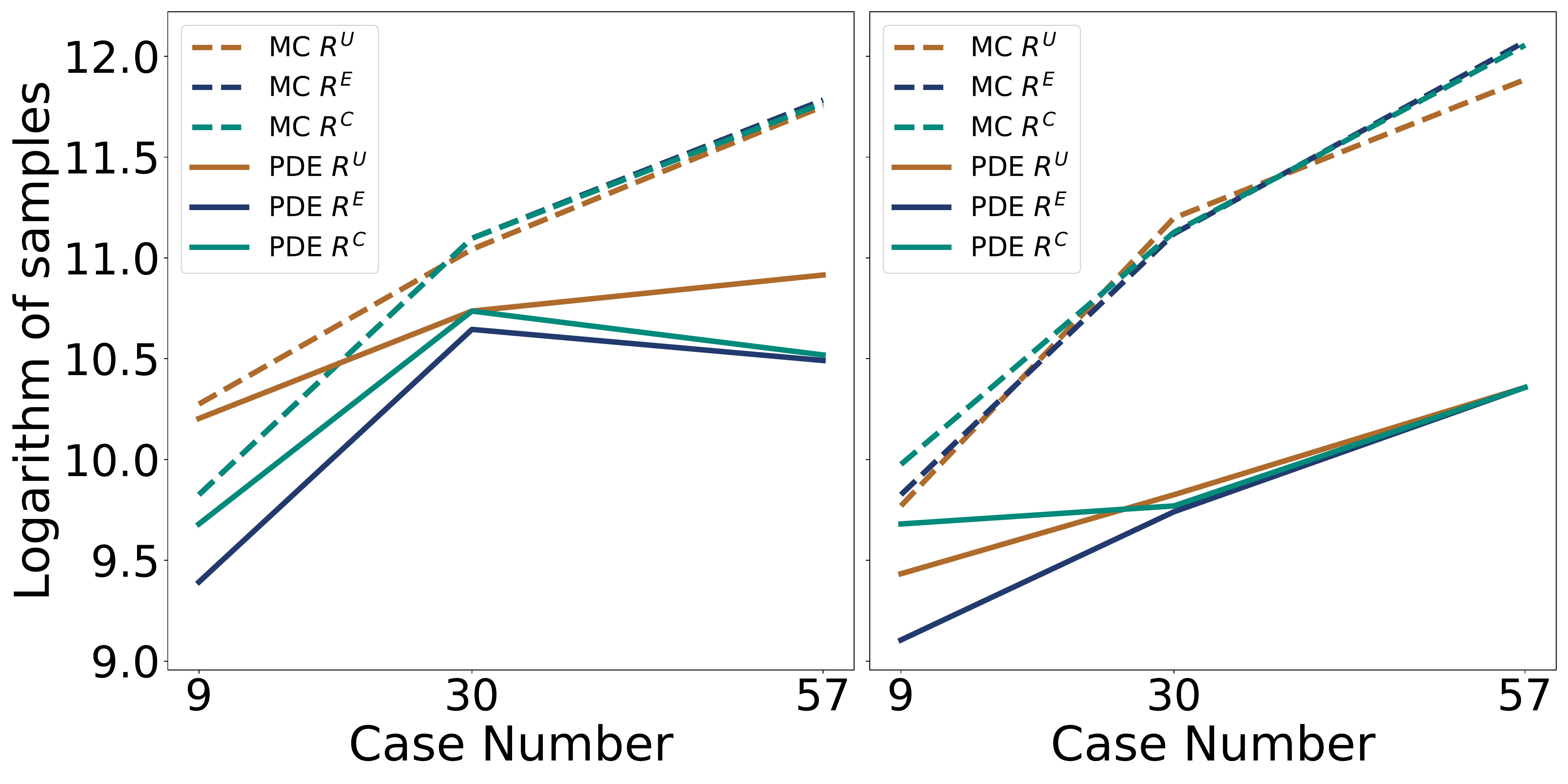}
	\caption{Total samples counts of the RO-PDF method (PDE) versus the MC with KDE method (MC) for all cases and noise correlations. Left: cases without line failures. Right: cases with line failures.}
	\label{fig:samples}
\end{figure}

For each test case, we calculate the ``true'' marginal PDFs of each speed (i.e., $f_{\w_i}(z;t)$ for all $i$) at discrete times by 30,000 MC simulations of~\eqref{mmc_modf} with Gaussian KDE and Silverman's rule. These PDFs are treated as yardsticks, and the number of simulations used to compute them was determined by a convergence study. To test the scalability of the RO-PDF method, we specify a 5\% error tolerance. For each speed, we then calculate the minimum number of MC simulations needed for its marginal PDF equation solution to be within the tolerance. Here, relative error is measured against the yardstick solution via the $L^2$ norm over space and time. We take the total (speed) sample count for our method to be the sum (over all speeds) of the number of MC simulations needed. We repeat this process for the MC method with KDE. That is, for every speed, we find the minimum number of simulations needed for the marginal PDF computed via KDE to be within 5\% relative error of the yardstick solution, and then sum over all speeds to get a total MC method sample count. Fig.~\ref{fig:samples} shows for all test cases, as the system's dimensionality increases, the total (speed) sample count, and therefore cost of MC with KDE increases considerably faster than that of the RO-PDF method. We omit the results for the angles because, in this case, the RO-PDF method outperforms MC with KDE by multiple orders of magnitude. All MC simulations of~\eqref{mmc_modf}, regardless of the method or use, were computed via Milstein's scheme with $\Delta t=0.005$. 

\section{Conclusion}
Most real-world power system models are inherently stochastic and high-dimensional, and the underlying stochastic noise processes can be correlated. Quantifying the uncertainty in such dynamics is challenging, and many, if not most methods suffer from the curse of dimensionality when it comes to characterizing PDFs of system states. The method of RO-PDF equations shows great promise in alleviating the curse of dimensionality for high-dimensional stochastic dynamical systems when it is sufficient to study the uncertainty of low-dimensional {\sc{QoI}}. We demonstrated this for scalar {\sc{QoI}} in the multi-machine classical model~\eqref{mmc_modf} with stochastic power injection driven by both correlated and uncorrelated OU processes. 

We plan to continue this work by considering larger systems (e.g., IEEE 300-BUS System, TAMU 2K, etc.) in order to more rigorously study the scalability of the method. Additionally, more complex/realistic models than~\eqref{mmc_modf}, such as those containing governor equations and detailed generator models, can be incorporated into the RO-PDF framework, as well as nonlinear and vector-valued {\sc{QoI}}. We are particularly interested in using this framework to quantify uncertainty of cascade failures. Lastly, real-world data can be used to achieve realistic estimates for the correlation between the underlying noise processes, i.e., the correlation matrix $\bm{R}$.


\bibliographystyle{ieeetr}
\bibliography{biblio}

\vfill

\end{document}